\documentclass[letterpaper, 10 pt, conference]{ieeeconf}  

\IEEEoverridecommandlockouts

\usepackage{cite}
\usepackage{amsmath,amssymb,amsfonts}
\usepackage{graphicx}
\usepackage{textcomp}
\usepackage{xcolor}
\usepackage{url}
\usepackage{hyperref}

\usepackage{tabularx}
\usepackage{array}

\newcolumntype{Y}{>{\raggedright\arraybackslash}X}

\usepackage[font=small]{caption}
\hypersetup{
colorlinks=true, 
urlcolor=cyan,
}

\usepackage[ruled,vlined,linesnumbered]{algorithm2e}

\usepackage{mathtools}
\usepackage{kpfonts}

\newcommand{\comment}[1]{}

\usepackage{tikz}
\usetikzlibrary{shapes.geometric, positioning, arrows.meta, calc, fit, quotes}

\usepackage{booktabs}
\usepackage{wrapfig}

\usepackage{listings}

\newcommand{\projectname}{GT-VLA}
\newcommand{\moe}{TA-MoE}
\newcommand{\ablationMoE}{GT-VLA-single}
\newcommand{\ablationTrace}{G-VLA}

\newcommand{\semanticpoint}[0]{g}
\newcommand{\tracepolicy}[0]{\pi^\tau}
\newcommand{\actpolicy}[0]{\pi^{\mathrm{act}}}
\newcommand{\planoracle}[0]{\Phi_{\mathrm{plan}}}
\newcommand{\targetoracle}[0]{\Phi_{\mathrm{target}}}
\newcommand{\overlay}[0]{\text{overlay}}

\def\BibTeX{{\rm B\kern-.05em{\sc i\kern-.025em b}\kern-.08em
    T\kern-.1667em\lower.7ex\hbox{E}\kern-.125emX}}

\begin{document}

\bstctlcite{BSTcontrol}


\title{\LARGE \bf
GT-VLA: Target-Conditioned Trace Guidance for Generalizable Robotic Manipulation\\

}

\author{
Ninghan Zhong$^{1*}$,
Jing-Chen Peng$^{1*}$,
Sriram Vishwanath$^{1}$%
\thanks{$^{*}$Equal contribution.}%
\thanks{$^{1}$School of Electrical and Computer Engineering,
Georgia Institute of Technology, Atlanta, GA, USA.
{\{nzhong34\}@gatech.edu}}%
}

\maketitle

\begin{abstract}
    Vision-Language-Action (VLA) models have shown strong performance on robotic manipulation, but they often struggle to generalize to unseen tasks, configurations, and long-horizon settings. A key challenge is that VLAs overfit to training scenes and fail to follow novel language instructions. Off-the-shelf vision-language models (VLMs) often provide stronger generalization, but cannot directly control robot actions. To combine the common sense of VLMs with VLA control, we propose \textbf{Guided Trace VLA} (\projectname), \textbf{a steerable framework that accepts guidance from an external generalist VLM through trace-conditioned action generation}.
    \projectname~uses a generalist model to identify semantic guidance for the current skill, converts this guidance into a 2D visual trace, and conditions its action policy on the resulting trace-rendered observation. This design separates semantic target acquisition, trace generation, and low-level action execution, allowing high-level guidance to propagate to robot actions. \projectname~uses a Mixture-of-Experts architecture with skill-specific trace and action modules for robust execution. We evaluate \projectname~on LIBERO and a physical robot platform, showing improved generalization over recent VLA baselines in both settings. The code and additional supplemental materials are available on our project website at {\footnotesize \url{https://ivaniz.github.io/gt-vla/}}.
\end{abstract}

\section{Introduction}
\label{sec:introduction}

\begin{figure*}[tb]
    \centering
    \includegraphics[width=\linewidth]{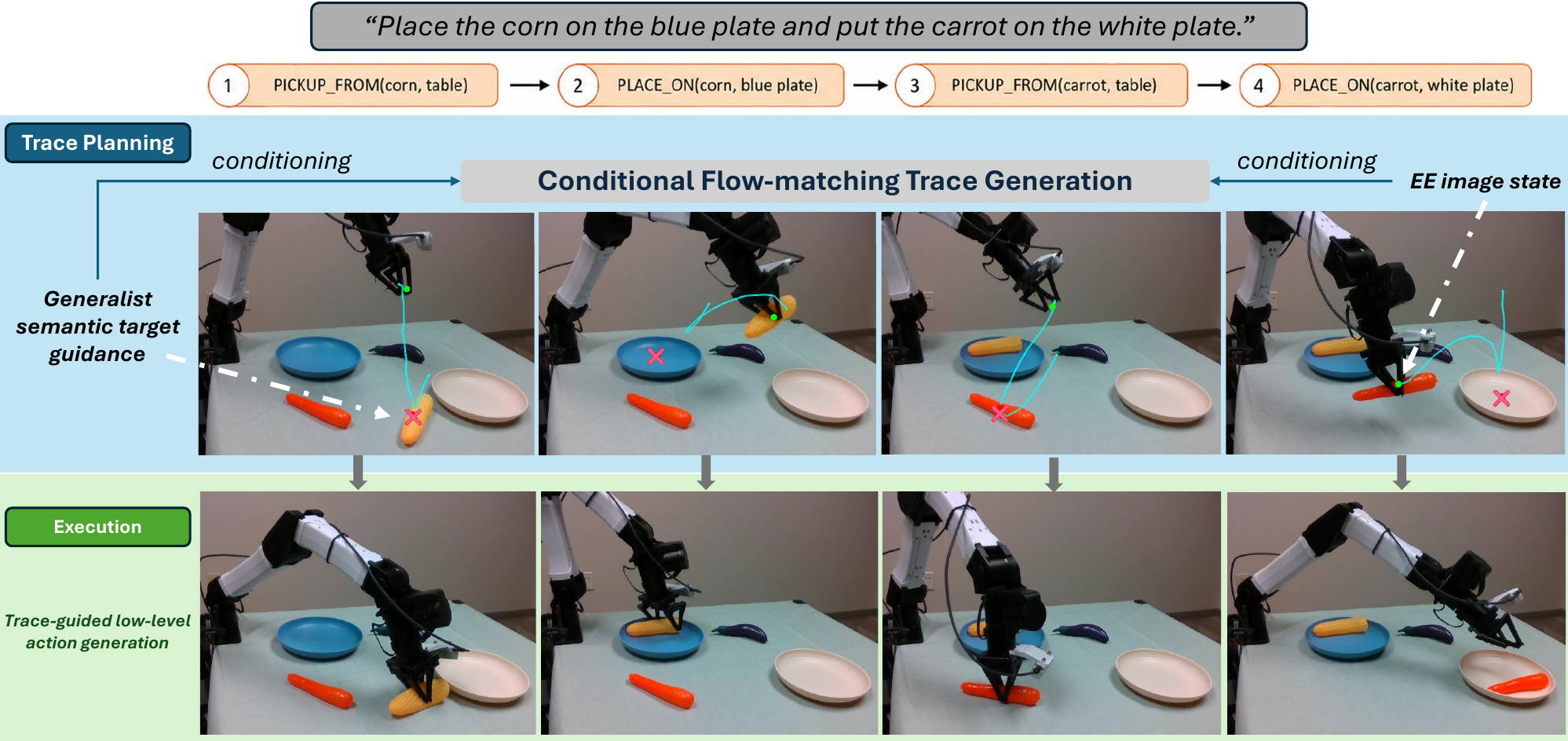}
    


    \vspace{-5pt}
    \caption{\projectname~overview. An off-the-shelf generalist VLM decomposes tasks into skills and provides skill-specific guidance.
    This is converted into an image-space trace conditioning for the action policy, ensuring execution follows high-level intent.}
    \label{fig:ame_overview}
    \vspace{-18pt}
\end{figure*}

Vision-Language-Action models (VLAs) integrate perception, planning, and control into unified policies, excelling at in-distribution manipulation \cite{black2024pi_0, pmlr-v270-kim25c}. However, they frequently suffer from scene overfitting, repeating memorized behaviors rather than adapting to novel instructions or unseen configurations \cite{fei2025liberoplusindepthrobustnessanalysis, zhan2026stablelanguageguidancevisionlanguageaction}. While decomposing complex tasks into atomic subtasks \cite{zhang2026atomicvlaunlockingpotentialatomic, Wu2025DoWY} or querying generalist VLMs for runtime steering \cite{Wu2025DoWY} improves robustness, these high-level semantic signals are often ignored by the low-level controller. Effective generalization requires a steerable execution pipeline that propagates external semantic guidance all the way to action generation.

To bridge this gap, we propose \textbf{Guided Trace VLA} (\projectname), a framework that injects guidance from an off-the-shelf generalist model through an intermediate, actionable visual representation.
\projectname~first queries a generalist VLM to decompose a task into a skill plan (Fig.~\ref{fig:ame_overview}).
For each skill, \projectname~generates 2D visual traces anchored to a semantic point predicted by the generalist VLM, which subsequently guides downstream action execution. 
This modularizes high-level intent and low-level control, preserving the VLA's competence while keeping it steerable and interpretable.

At the core of \projectname~is a Trace-Action Mixture-of-Experts (\moe) architecture featuring skill-specific trace and action heads. The trace head uses conditional flow-matching to generate traces conditioned on the semantic target.
This trace is then used to condition action generation.
We further propose a set of data augmentations to encourage the model to respect semantic guidance and be robust to imperfect generated traces.

We evaluate \projectname~in simulation and the real world. In simulation, we use a clean LIBERO train-test split to test generalization.
We also validate \projectname~on a physical robot platform.
Across both settings, \projectname~outperforms recent VLA baselines, demonstrating the benefit of semantic-trace conditioning for steerable and generalizable manipulation.

Our core contributions are:
\begin{itemize}
\item \textbf{\projectname}: A framework that translates high-level VLM guidance into visual traces to steer low-level VLA execution on unseen tasks and configurations.
\item \textbf{Trace-Action Mixture-of-Experts (\moe)}: A modular architecture coupling skill-specific trace generation with trace-guided action execution.
\item \textbf{Robust Trace-Conditioning Strategies}: Guidance-conditioned generation, trace perturbation, and scene masking ensure the policy respects guidance while remaining robust to imperfect traces.
\end{itemize}
\section{Related Work}
\label{sec:related_work}

\subsection{Vision-Language-Action Models}

VLAs leverage large-scale pre-training and semantic priors to unify language understanding, visual perception, planning, and control within a single robotic policy~\cite{pmlr-v270-kim25c, black2024pi_0}.
Recent VLAs achieve improved performance across diverse tasks~\cite{black2025pi_, zitkovich2023rt}.
However, they still struggle with unseen objects, novel scene configurations, and long-horizon task compositions~\cite{fei2025liberoplusindepthrobustnessanalysis}.

To improve long-horizon execution, some methods decompose tasks into short subtasks, using embodied chain-of-thought to switch between reasoning and acting~\cite{%
Wu2025DoWY
} or directly predicting an atomic skill to execute~\cite{zhang2026atomicvlaunlockingpotentialatomic}. 
Although these methods improve subtask composition, VLAs still overfit to visual observations, reproducing memorized behaviors even when instructed otherwise~\cite{zhan2026stablelanguageguidancevisionlanguageaction}.

Another line of work uses explicit high-level reasoners to generate plans or intermediate decisions that are executed step by step by a lower-level VLA~\cite{intelligence2026pi07steerablegeneralistrobotic, zhou2025chatvla, black2025pi_, zawalski2025roboticcontrolembodiedchainofthought}. These hierarchical designs, introduced in embodied chain-of-thought policies~\cite{zawalski2025roboticcontrolembodiedchainofthought}, improve generalization by separating semantic reasoning from control. Some methods freeze the VLM backbone to preserve general-purpose reasoning, but this may reduce task performance~\cite{black2025pi_, zhou2025chatvla}. Our work builds on the view that hierarchy is useful but insufficient: effective generalization requires a framework that can propagate guidance to the action generation level.

\subsection{Runtime Verification and Steering}

Runtime verification and steering improve VLA robustness by introducing inference-time signals that verify or guide robot actions. Learned critics can steer candidate actions~\cite{nakamoto2024steering}, but may not generalize to novel scenes or tasks. Mechanistic interpretability methods can control fine-grained behavior attributes such as speed or clearance~\cite{haon2025mechinterp}, while ReSteer~\cite{chen2026resteerquantifyingrefiningsteerability} evaluates and improves steerability with targeted data collection.

Recent methods also use VLMs for runtime verification and guidance. RoboMonkey~\cite{pmlr-v305-kwok25a} selects among sampled actions using a fine-tuned VLM, LoHo-Manip~\cite{liu2026long} fine-tunes a VLM for task decomposition and visual guidance, and AHA~\cite{duan2024aha} trains a VLM to detect and explain manipulation failures.
FOREWARN~\cite{wu2025foresightforethoughtvlminthelooppolicy} 
aligns a VLM with latent-space predictions to steer robot actions.
These methods show the promise of VLM-based steering, but often require additional fine-tuning of the high-level VLM.

Off-the-shelf VLMs can also directly generate policies~\cite{liangCodePolicy}, orchestrate tools~\cite{shi2025maestroorchestratingroboticsmodules}, or act as performance critics~\cite{
Wu2025DoWY}. However, such external verification cannot correct execution if the underlying VLA fails to incorporate the guidance.
Our work follows the runtime steering approach, using an off-the-shelf generalist model for semantic guidance and designing the downstream pipeline to be steerable by this guidance.

\subsection{Visual Prompting and Trace Guidance}

Visual prompting can provide spatial guidance to generalist models. In computer vision, the SAM series shows that simple geometric point and box prompts can guide zero-shot segmentation~\cite{%
carion2025sam
}. Related robotics methods use visual hints or intermediate representations to connect semantic intent with low-level action. VLMs can generate semantic guidance and constraints for motion primitive execution~\cite{zhu2025bridgingvlmkmpenabling};
MOKA~\cite{fangandliu2024moka} uses keypoint affordances from a VLM to command a low-level VLA;
SpatialVLA~\cite{qu2025spatialvlaexploringspatialrepresentations} studies 3D spatial encodings of action tokens; 
\textit{Fast-ThinkAct}~\cite{%
huang2026fast} connects semantic guidance to control with visual latent planning.

A growing line of work uses traces as visual guidance for manipulation. 
TraceVLA~\cite{zheng2025tracevla} uses historical end-effector traces to improve spatial-temporal awareness, while RT-Trajectory~\cite{gu2023rttrajectory} shows that trajectory sketches can act as steerable policy inputs. 
ATM~\cite{wen_lin_so_chen_dou_gao_abbeel_2024} predicts motion flow for the entire scene. 
Closer to our setting, LoHo-Manip~\cite{liu2026long} uses a fine-tuned manager for subtask decomposition and image-space trace planning, followed by a trace-conditioned executor, while MolmoAct~\cite{molmoact2025} jointly predicts depth, traces, and actions within a monolithic action-reasoning model. 
\projectname~differs from prior work by both explicitly decomposing long-horizon tasks into skills, and accepting guidance from an external, off-the-shelf generalist VLM without fine-tuning, decoupling semantic guidance from trace generation and low-level control.
\begin{figure}[htb]
    \centering
    \includegraphics[width=\linewidth]{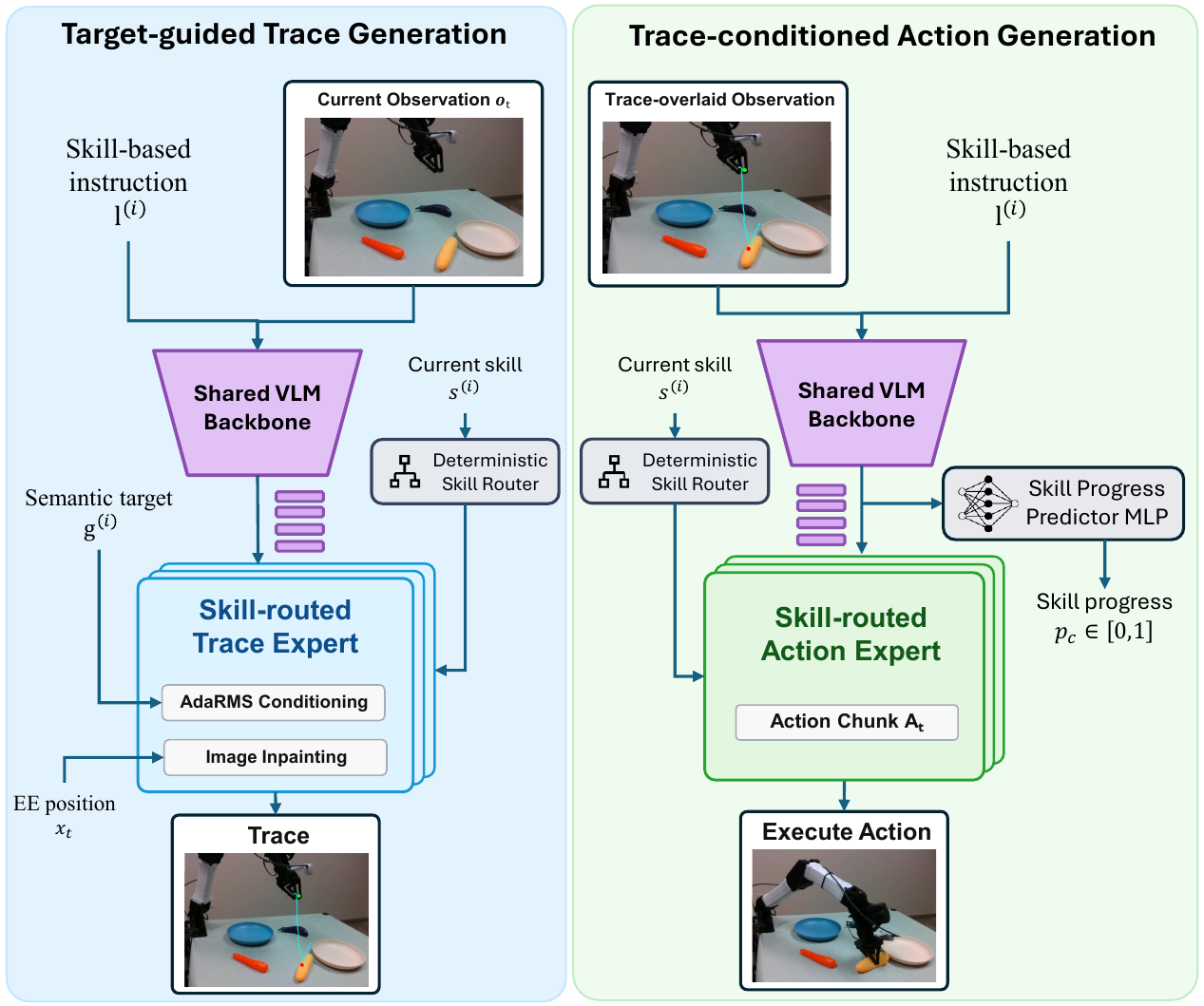}
    \vspace{-15pt}
    \caption{\projectname~architecture. \textbf{Left}: a skill-routed trace expert
generates a 2D trace anchored to the semantic target $g^{(i)}$ and end-effector position $x_t$. \textbf{Right}: a
skill-routed action expert predicts actions from the trace-overlaid
observation, and an MLP predicts skill progress. A deterministic
router selects experts for the current skill.}
    \label{fig:system_overview}
\end{figure}

\section{Method}
\label{sec:method}

The central goal of \projectname~is to make VLA execution steerable by semantic guidance from an off-the-shelf generalist VLM.
Instead of relying on the VLA to infer all of the task semantics, \projectname~uses an off-the-shelf generalist VLM to provide sparse guidance at inference time.
This guidance is converted into a dense visual trace to aid action generation, injecting high-level intent while preserving the VLA's learned low-level control competence.

Given an instruction $\ell$ and observation $o_t$, the generalist VLM first decomposes the task into parameterized manipulation skills. At the start of each skill, the same generalist model identifies a semantic target point in the image, such as an object to grasp, a handle to manipulate, or a placement location. A skill-specific trace expert then generates a 2D end-effector trace conditioned on this target, and the corresponding action expert executes low-level actions given the trace and observation. 
Fig.~\ref{fig:system_overview} illustrates the architecture, and Alg.~\ref{alg:aimvla-inference} gives the inference pseudocode.

\subsection{Skill Interface and Semantic Target Acquisition}
\label{sec:skill_inference}

We use a finite library of manipulation skills $\mathcal{S}$, where each skill $s \in \mathcal{S}$ is represented as a semi-structured operator with arguments.
For example, the \textit{PLACE} skill may be instantiated as \textit{PLACE(bowl, table)}.
Given a task instruction $\ell$ and observation $o_t$ containing camera views and robot state, we query an off-the-shelf generalist VLM $\Phi$ to produce a skill plan
\begin{equation}
\label{eq:skill_plan}
P = \planoracle(\ell, o_t)
= \left(s^{(1)}, s^{(2)}, \ldots, s^{(k)}\right),
\end{equation}
where each $s^{(i)} \in \mathcal{S}$ is an atomic skill to be executed in sequence.
This decomposition reduces a long-horizon instruction into shorter manipulation segments, but does not by itself guarantee that the low-level policy will execute each segment in the intended way.
To make skill execution steerable,
\projectname~additionally queries the generalist model for a semantic target point at the start of each skill. For the current skill $s^{(i)}$, the generalist model predicts a target in normalized image coordinates:
\begin{equation}
\label{eq:semantic_target}
g^{(i)} = \targetoracle(\ell, o_t, s^{(i)}) \in [0,1]^2,
\end{equation}
$g^{(i)}$ represents a spatial goal for the current skill.
For example, it may mark the 
object to pick up, the handle of a drawer to open, or the desired placement location.
This target serves as a semantic anchor for trace generation, allowing the policy to convert sparse high-level guidance signal into a feasible end-effector path.

\subsection{Target-Conditioned Trace Generation}
\label{sec:method-trace-gen}
The purpose of the trace generator is to translate the sparse semantic target into an actionable visual guidance signal.
Let $x_t \in [0,1]^2$ denote the projection of the robot end-effector position into image coordinates at timestep $t$.
We define the ground-truth remaining trace as the future portion of the image-space trajectory that completes the skill:
\begin{equation}
\label{eq:trace_gt}
\tau_t^* = \left(x_t, x_{t+1}, \ldots, x_{t^{\mathrm{end}}}\right),
\end{equation}
where $t^{\mathrm{end}}$ is the final timestep of the current skill.
In practice, we resample each trace by linear interpolation along its arc length to obtain a fixed-length sequence $\bar{\tau}_t^* \in \mathbb{R}^{H \times 2}$ where $H$ is the trace length.
We predict this semantic trace using a trace prediction model $\tracepolicy$:
\begin{equation}
\label{eq:trace_dist}
\tracepolicy\left(\tau_t \mid o_t, x_t, g^{(i)}, \ell^{(i)}\right).
\end{equation}
The language input $\ell^{(i)}$ combines $\ell$ and $s^{(i)}$, giving both high-level and skill-specific information.
$\pi^{\tau}$ is implemented as a flow-matching model, with the loss
\begin{equation}
\label{eq:trace_loss}
\begin{split}
\mathcal{L}^{\mathrm{trace}}
& = \mathbb{E}_{(\bar{\tau}_t^*, o_t, \ell^{(i)}, g^{(i)}) \sim \hat{\mathcal{D}}, \epsilon, \lambda}
\left[
\left\|
\hat{v}
-
(\epsilon - \bar{\tau}_t^*)
\right\|_2^2
\right], \\
\hat{v} & \equiv v_\theta^{\tau}(\bar{\tau}_t^*(1-\lambda)+\epsilon\lambda, \lambda, o_t, x_t, g^{(i)}, \ell^{(i)})
\end{split}
\end{equation}
where $\hat{\mathcal{D}}$ is a skill-annotated demonstration dataset, $\lambda \in [0, 1]$ is the flow-matching time, $\epsilon$ is the noise vector, and $v_\theta^{\tau}$ is the predicted trace velocity field to be learned.
We draw $\lambda$ and $\epsilon$ following \cite{black2025pi_}.
Ground-truth traces are obtained by projecting the end-effector position into the camera image. Details on dataset construction are provided on the project website.

The current end-effector position $x_t$ is incorporated as an inpainting-style constraint to clamp the trace starting point. 
%
%
The semantic target provides a second constraint, injected through Adaptive RMSNorm conditioning (Fig.~\ref{fig:system_overview} left). 
These constraints enforce the trace to start from the robot's current image-space position and remain anchored to the spatial goal.

\subsection{Robust Trace-Conditioned Action Learning}
\label{sec:action-learning}
\comment{
}
The trace provides an interpretable 2D plan, but the action expert must map it to 6-DoF robot motion and gripper commands. We train the policy to follow trace guidance while remaining robust to imperfect traces and 2D--3D ambiguity.
The action expert is implemented as a flow-matching model following~\cite{black2024pi_0}, with the trace $\tau_t$ rendered onto the visual observation:
\begin{equation}
\label{eq:action_dist}
\actpolicy\left(A_t \mid \overlay(o_t, \tau_t), \ell^{(i)}\right).
\end{equation}

\begin{figure}[htbp]
    \centering
    \includegraphics[width=\linewidth]{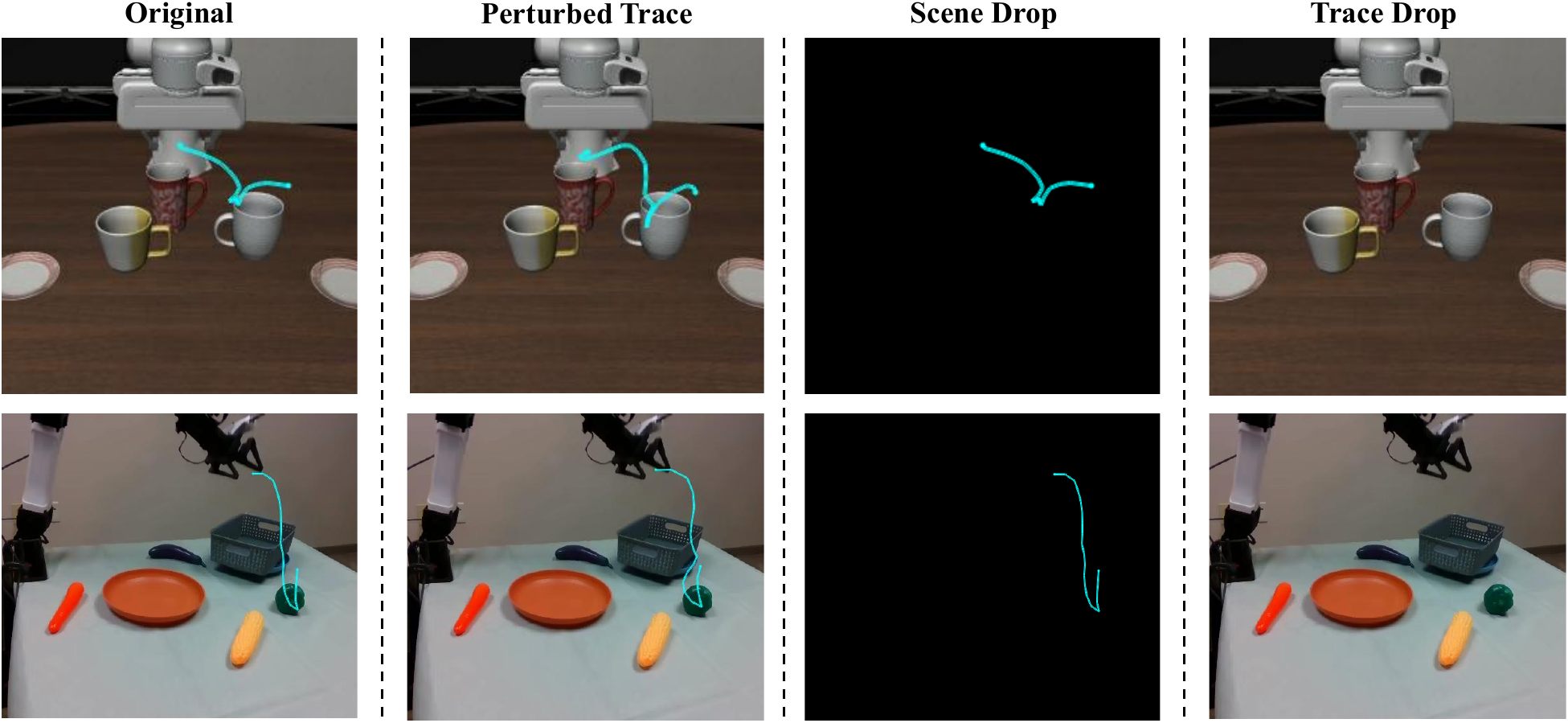}
    \vspace{-10pt}
    \caption{Trace-conditioning augmentations (Sec.~\ref{sec:action-learning}),
    shown on LIBERO (top row) and the real robot (bottom row). 
    }
    \label{fig:data_aug}
\end{figure}

To force the action expert to respect the trace guidance while considering scene geometries, and to bridge the gap between training demonstrations and imperfect inference traces, we propose three data augmentation techniques.
Visual illustrations are provided in Fig.~\ref{fig:data_aug}.

\textbf{Random scene drop.}
To encourage the action expert to respect the trace, we randomly mask the scene observation while keeping the rendered trace visible. This forces the policy to sometimes act primarily from the trace guidance, reducing its tendency to overfit to memorized visual configurations in the training scenes.

\textbf{Random trace drop.}
Because the trace is defined in 2D image space, blindly following it can be harmful when depth, occlusion, or contact geometry matters.
We randomly remove the trace overlay during training, providing only the clean image. This preserves the policy's ability to reason from the original scene image and robot state, reducing the learned controller's sensitivity to inaccurate or geometrically ambiguous traces.

\textbf{Low-frequency trace perturbation.}
To bridge the distribution gap between ground-truth traces at training time and generated traces at inference time, we perturb the rendered training trace with smooth low-frequency noise.
This produces traces that remain plausible end-effector paths but are slightly shifted or bent, addressing distribution shift between the training traces and generated traces at inference time.

\subsection{Trace-Action Mixture-of-Experts Architecture}

\label{sec:architecture}
Following the recipe from \cite{black2024pi_0}, we instantiate $\tracepolicy$ and $\actpolicy$ as small ``expert" transformers attending to a shared VLM backbone.
We instantiate the VLM backbone using a pretrained checkpoint of Gemma 2B \cite{black2025pi_}.

One potential challenge of image-trace conditioning is ambiguity: \textit{PICK} and \textit{PLACE} skills have similar traces but require different actions.
Inspired by \cite{zhang2026atomicvlaunlockingpotentialatomic}, we instantiate $\actpolicy$ as a deterministically-routed model
with one expert assigned to each skill category in our dataset.
We refer to this as a skill-routed mixture-of-experts (MoE) following~\cite{zhang2026atomicvlaunlockingpotentialatomic}.
At training and inference time, only the expert corresponding to the skill being run is invoked.
Each expert is a 300M parameter transformer, following the action expert design of \cite{black2024pi_0}.

We implement $\tracepolicy$ as a lightweight flow-matching transformer ($\sim$60M parameters), following the same skill-routed MoE design as the action generator.

We also attach a lightweight MLP head (Fig.~\ref{fig:system_overview} right) to the final-layer hidden states of the shared VLM backbone, which predicts the progress $p_c \in [0,1]$ of the current skill. At inference, GT-VLA advances to the next skill once $p_c \geq \eta_c$ (Alg.~\ref{alg:aimvla-inference}), where $\eta_c$ is a chosen threshold.
This controls transitions between steps during inference.

\begin{figure*}[ht]
    \centering
    \includegraphics[width=\linewidth]{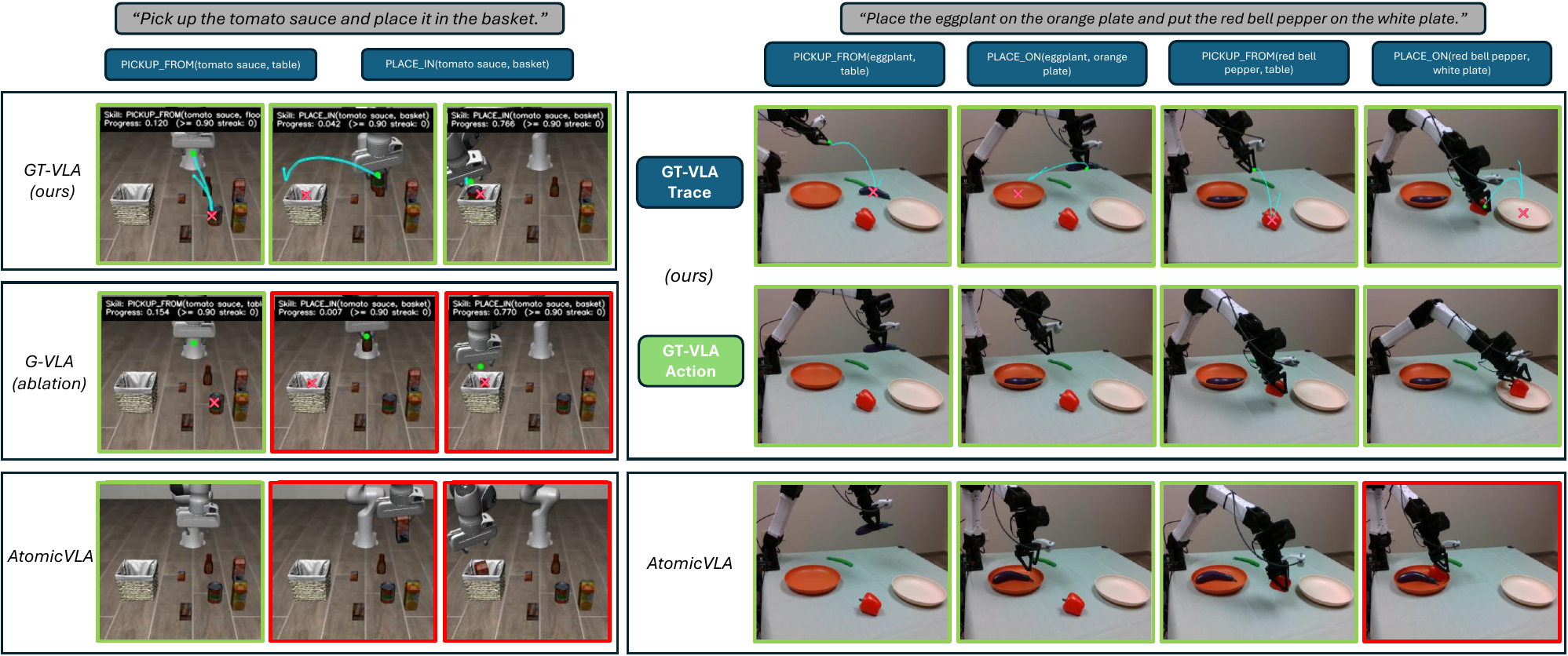}
    \vspace{-15pt}
    \caption{\textbf{Comparing steerability of \projectname~against baselines and ablations.}
    On the left we show a task from Libero-Object, which is an OOD task. 
    \projectname~(top-left) is steerable by the external semantic target.
    \ablationTrace~executes memorized actions despite having correct semantic guidance (mid-left).
    AtomicVLA (bottom-left) also fails to generalize to this instruction.
    In real-world experiments, we show a 4-step OOD task.
    \projectname~completes all skills, while AtomicVLA fails to adapt.}
    \label{fig:running_demos}
    \vspace{-15pt}
\end{figure*}

\subsection{Inference Algorithm}
\label{app:inference_alg}

\SetKwComment{Comment}{// }{}
\newcommand{\mycommfont}[1]{%
    {\footnotesize\textcolor{gray}{#1}}%
}

\newlength{\algcommentgap}
\setlength{\algcommentgap}{0.05em}

\newcommand{\RComment}[1]{%
    \unskip\hspace{\algcommentgap}%
    \Comment*[f]{#1}\par%
}

\SetCommentSty{mycommfont}
\SetKw{Continue}{continue}
\SetKw{Break}{break}

{
\begin{algorithm}[htbp]
    \small
    \DontPrintSemicolon
    \SetInd{0.3em}{0.6em}  
    \caption{GT-VLA Inference} 
    \label{alg:aimvla-inference}
    \KwIn{$\ell$, $o_{\text{init}}$, $f_{\text{upd}} \geq 1$, $\eta_{\mathrm{c}} \in [0, 1]$}

    $o \gets o_{\text{init}}$   \Comment*[r]{initialize obs.}
    
    $\{s^{(1)}, \dots, s^{(k)}\} \gets \planoracle(\ell, o)$ \Comment*[r]{get skill plan (Sec.~\ref{sec:skill_inference})}

    \For{each skill $s^{(i)} \in \{s^{(1)}, \dots, s^{(k)}\}$} {
    
        $\ell^{(i)} \gets \text{concat}(\ell, s^{(i)})$ \Comment*[r]{concat prompt and skill}
        
        $\semanticpoint^{(i)} \gets \targetoracle(\ell, o, s^{(i)})$ \Comment*[r]{semantic target (Sec.~\ref{sec:skill_inference})}

        $\textup{skillDone} \gets \text{False}$\;

        \While{\textup{not} \textup{skillDone}}{
        
            $x \gets \text{project}(o)$ \Comment*[r]{end effector image projection}

            $\tau \sim \tracepolicy(o, x, \semanticpoint^{(i)}, \ell^{(i)})$ \Comment*[r]{generate trace (Sec.~\ref{sec:method-trace-gen})}

            \For{each $j = 1, \ldots, f_{\text{upd}}$}{

                $\tilde{o} \gets \overlay(o, \tau)$ \Comment*[r]{overlay visual trace}

                $p_{\mathrm{c}} \gets \text{comp}(\tilde{o},\ell^{(i)})$ \Comment*[r]{skill progress (Sec.~\ref{sec:architecture})}

                \If{$p_{\mathrm{c}} \geq \eta_{\mathrm{c}}$} {
                    $\textup{skillDone} \gets \textup{True}$  \;
                    \Break\;
                }

                $A \sim \actpolicy(\tilde{o}, \ell^{(i)})$  \Comment*[r]{sample actions (Sec.~\ref{sec:action-learning})}

                $o \gets \text{execute}(A)$ \Comment*[r]{execute actions, update obs.}
                
            }
            
        }
    }
\end{algorithm}
}

\projectname's inference pipeline is presented in Algorithm~\ref{alg:aimvla-inference}. 
The algorithm takes as input a natural language prompt $\ell$, current observation $o_t$, a trace update frequency $f_{\text{upd}} \geq 1$ that controls the number of action expert executions between each trace refresh, and a skill completion threshold $\eta_{\mathrm{c}} \in [0, 1]$.
First, \projectname~queries a generalist VLM to decompose the task into a skill plan (Line 2).
For each skill, the algorithm constructs a structured skill command and queries the VLM for a semantic target (Lines 4, 5).
Execution then proceeds via a nested loop. The outer loop regenerates a visual trace every $f_{\text{upd}}$ action chunks (Lines 8, 9). The inner loop first obtains the trace overlaid visual observation (Line 11) and predicts the current skill progress $p_{\mathrm{c}} \in [0, 1]$ (Line 12). If the predicted progress reaches the completion threshold $\eta_{\mathrm{c}}$ (Lines 13-15), the algorithm proceeds to the next skill. Otherwise, the algorithm samples and executes actions, and updates current observation (Lines 16, 17).

In our current formulation, the high-level skill plan is executed open-loop.
Algorithm~\ref{alg:aimvla-inference} can be modified to execute the skill plan in a closed-loop manner, where the high-level generalist VLM is periodically queried to replan the skill sequence to improve failure recovery. 
We leave evaluation of this extension to future work.

\section{Experiments}
\label{sec:experiments}

Our evaluation involves four major components: (1) comparisons against recent baselines
(Sec.~\ref{sec:sim-setup}--\ref{sec:real-world-eval-results}),
(2) ablation studies on proposed designs (Sec.~\ref{sec:ablation}), (3) pilot studies on \projectname's robustness to guidance noise and generalist VLM modularity (Sec.~\ref{sec:noise-benchmark}, \ref{sec:eval-vlm-swap}), and finally (4) failure analysis of \projectname~(Sec.~\ref{sec:failure-analysis}).

\emph{Comparisons against Recent Baselines:} We evaluate \projectname~in both simulation and on a physical robot platform. In this part, our experiments are designed to answer two questions. First, does \projectname~improve generalization to unseen tasks, object configurations, and task compositions compared with recent VLA baselines? 
Second, does converting semantic guidance from an off-the-shelf VLM into an explicit and trace-conditioned execution signal improve steerability beyond directly relying on the generalist model for steering or verification without explicit design integrations?

We compare \projectname~against three baselines with publicly released
implementations. The base $\pi_{0.5}$ model~\cite{black2025pi_}, on which our method and the other baselines are built, directly maps vision-language inputs to actions.
AtomicVLA~\cite{zhang2026atomicvlaunlockingpotentialatomic} predicts which skill is being executed at every step, and uses a MoE to learn each skill.
SEAL~\cite{Wu2025DoWY} is a VLA steering framework using an external VLM verifier.
We use Gemini 3.1 Pro to implement both SEAL and \projectname.

We additionally provide a best-effort reproduction of LoHo-Manip~\cite{liu2026long} on our project website; since no official implementation is available, we report it as a supplementary comparison.

\emph{Ablation Studies: } We conduct ablation studies to evaluate two core design choices in \projectname: the use of a trace-generation stage, the \moe~architecture, and the three data augmentation techniques for trace-conditioned learning. Fig.~\ref{fig:running_demos} illustrates \projectname's steerability, showing how \projectname~adapts to unseen configurations compared to baseline and ablation methods.

\subsection{Simulation Setup}
\label{sec:sim-setup}

We evaluate \projectname~on LIBERO~\cite{liu2023libero}. To evaluate generalizability to unseen tasks, we forgo per-suite in-distribution finetuning. Instead, we use a clean train-test split where methods are trained on Libero-10 and Libero-90 demonstrations~\cite{Wu2025DoWY} and evaluated on the Libero-Goal, Libero-Spatial, and Libero-Object suites~\cite{liu2023libero}.
We preprocess the dataset to obtain task decompositions, semantic targets, and ground truth traces.
The annotation procedure is similar to approaches from recent work~\cite{Wu2025DoWY, zhang2026atomicvlaunlockingpotentialatomic}, with details on our project website.
We train all the methods for 100k iterations with batch size 64 on 4 GH200 GPUs.

\subsection{Simulation Results}
\label{sec:sim-eval-results}

Table~\ref{tab:sim-eval-results} reports the simulation results.
\projectname~achieves the best performance across all three LIBERO evaluation suites, with an average success rate of 43.8\%. This corresponds to a 28.0 percentage point improvement over vanilla $\pi_{0.5}$, a 9.9 point improvement over AtomicVLA, and a 16.5 point improvement over SEAL.

Compared to \projectname, AtomicVLA also uses a skill-routed MoE to handle different skills, but does not support high-level generalist guidance or trace conditioning.
The stronger performance of \projectname, particularly on LIBERO-Object, suggests that generalist information and trace guidance help performance when faced with novel objects.
%
Although both SEAL and \projectname~use an external generalist VLM, \projectname~achieves substantially higher performance.
This suggests that external guidance is not sufficient for steering: The VLA must also be designed to accept guidance effectively. 

\begin{table}[t]
\caption{Success rate on LIBERO out-of-distribution evaluation suites, with 50 trials $\times$ 10 tasks per category (\%). 95\% Wilson confidence intervals are shown in parentheses.}
\vspace{-5pt}
\label{tab:sim-eval-results}
\centering
\scriptsize
\setlength{\tabcolsep}{2.5pt}
\begin{tabular}{lcccc}
\toprule
\textbf{Models} & \textbf{Goal} & \textbf{Spatial} & \textbf{Object} & \textbf{Avg.} \\
\midrule
$\pi_{0.5}$         & 16.2 (13.1--19.7) & 29.6 (25.6--33.8) & 1.6 (0.7--3.1) & 15.8 \\
AtomicVLA           & 30.2 (26.2--34.4) & 46.8 (42.4--51.3) & 24.6 (20.9--28.6) & 33.9 \\
SEAL                & 31.0 (27.0--35.3) & 45.0 (40.6--49.5) & 6.0 (4.1--8.5) & 27.3 \\
\projectname~(ours) & \textbf{38.8 (34.5--43.2)} & \textbf{50.8 (46.3--55.3)} & \textbf{41.8 (37.4--46.3)} & \textbf{43.8} \\
\bottomrule
\end{tabular}
\end{table}

\subsection{Real-world Experiment Setup}
\label{sec:real-world-setup}

We further evaluate \projectname~on a physical tabletop manipulation platform. The setup consists of a YAM robot arm equipped with side-view and wrist cameras. The workspace contains plates, baskets, and fake vegetables with different colors, shapes, and sizes. The tasks are long-horizon pick-and-place instructions such as \textit{``place the red bell pepper into the green plate''}, typically involving two to six \texttt{PICK} and \texttt{PLACE} skills.

For real-world experiments, we compare against $\pi_{0.5}$ and AtomicVLA. We do not include SEAL because it requires a high-fidelity world model or digital twin for forward simulation, which is not available for our hardware setup.
We collect 299 complete episodes for training, each containing two to six \texttt{PICK} and \texttt{PLACE} skills. 
%
Task definitions and dataset annotations are detailed on our project website.
All methods are trained for 30k iterations with batch size 64 on 4 GH200 GPUs.
Inference is performed on a single GH200 GPU.

To evaluate real-world generalization, we construct five evaluation suites. The first two are \textbf{composition} suites, where each task is composed of skill configurations that appear in training, but arranged in unseen orders.
This tests whether the model can compose familiar behaviors into new long-horizon instructions.
We evaluate 4- and 6-skill variants, denoted as \textbf{composition-4} and \textbf{composition-6}.
The other three are \textbf{OOD} suites, where the skill configurations themselves are unseen.
We evaluate 2-, 4-, and 6-skill variants, denoted \textbf{OOD-2}, \textbf{OOD-4}, and \textbf{OOD-6}, respectively.
Each suite contains 10 tasks.
For each task, we run 5 trials with different initial object placements, resulting in 50 trials for each method-suite pair. We report episode-level success: A trial is successful only if the robot completes the full task.

\subsection{Real-world Experiment Results}
\label{sec:real-world-eval-results}

\begin{figure}[t]
    \centering
    \includegraphics[width=\linewidth]{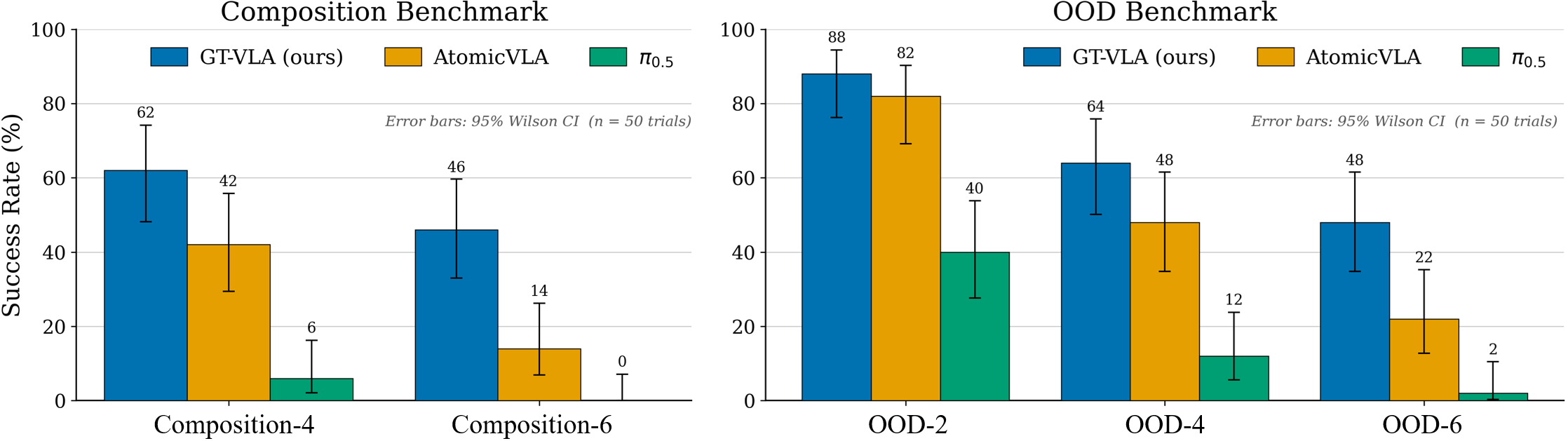}
    \caption{Real-world results. 5 trials $\times$ 10 tasks per category.}
    \label{fig:hardware_exp_results}
    \vspace{-7pt}
\end{figure}

Fig.~\ref{fig:hardware_exp_results} shows our real-world experiment results. \projectname~outperforms baselines across 5 real-world suites, showing target-conditioned trace guidance's effectiveness in real settings. \projectname's advantage over AtomicVLA widens in both the composition and OOD settings as task horizon increases, supporting our hypothesis that complex manipulation requires both continuous semantic alignment and subtask decomposition.

\subsection{Ablation Studies}
\label{sec:ablation}
We perform five ablation studies to isolate the contribution of the key components in \projectname: the trace generation stage, the MoE architecture, and the three data augmentation techniques. All ablations are evaluated in simulation on LIBERO following Sec.~\ref{sec:sim-setup}.

The \textit{\ablationTrace}~removes the trace module while keeping the semantic target guidance. We retain the same MoE action architecture and inject the target directly into the action heads using AdaRMS conditioning. This tests if converting the target into an explicit trace can provide additional benefit over using the target directly. The \textit{\ablationMoE}~removes the skill-routed MoE, using only a single trace-generation head and a single action head. This tests the impact of using skill-specific experts to model each skill separately.

To evaluate the contributions from the three data augmentation techniques for trace-conditioned learning, we train a leave-one-out ablation for each augmentation. \textit{No S. Drop} removes the random scene drop, \textit{No T. Drop} removes the random trace drop, and \textit{No Perturb.} removes the low-frequency trace perturbation.

\begin{table}[t]
\caption{Ablation results on LIBERO (\%). Results use 50 trials $\times$ 10 tasks per category, except the generalist VLM swap rows, which use 20 trials $\times$ 10 tasks. Parentheses show 95\% Wilson confidence intervals.}
\vspace{-5pt}
\label{tab:ablation}
\centering
\scriptsize
\setlength{\tabcolsep}{2.5pt}
\begin{tabular}{lcccc}
\toprule
\textbf{Models} & \textbf{LIBERO Goal} & \textbf{LIBERO Spatial} & \textbf{LIBERO Object} & \textbf{Avg.} \\
\midrule
\ablationMoE
& 27.6 (23.7--31.7)
& 40.6 (36.3--45.0)
& 27.8 (23.9--32.0)
& 32.0 \\

\ablationTrace
& 32.2 (28.1--36.5)
& 39.6 (35.3--44.0)
& 20.8 (17.3--24.6)
& 30.9 \\

No S. Drop
& 38.4 (34.2--42.7)
& 29.0 (25.2--33.1)
& 24.4 (20.8--28.4)
& 30.6 \\

No T. Drop
& 36.6 (32.5--40.9)
& 41.8 (37.6--46.2)
& 21.4 (18.0--25.2)
& 33.3 \\

No Perturb.
& 34.8 (30.8--39.1)
& 40.6 (36.4--45.0)
& 23.8 (20.3--27.7)
& 33.1 \\
\midrule

GPT-5.6
& 33.0 (26.9--39.8)
& 44.5 (37.8--51.4)
& 35.0 (28.7--41.8)
& 37.5 \\

Qwen 3.8
& 35.0 (28.7--41.8)
& 37.0 (30.6--43.9)
& 30.5 (24.5--37.2)
& 34.2 \\

Gemini 3.6
& 31.0 (25.0--37.7)
& 46.5 (39.7--53.4)
& 33.5 (27.3--40.3)
& 37.0 \\
\midrule

\projectname~(ours)
& \textbf{38.8 (34.5--43.2)}
& \textbf{50.8 (46.3--55.3)}
& \textbf{41.8 (37.4--46.3)}
& \textbf{43.8} \\
\bottomrule
\end{tabular}
\end{table}

Table~\ref{tab:ablation} reports the ablation results.
\projectname~achieves substantially higher success rates compared to \ablationTrace.
This suggests that directly injecting a semantic target into the action model is not enough, and the target is much more useful if it is converted into an actionable trace that the policy can visually follow. This is also visualized in Fig.~\ref{fig:running_demos} (mid-left) where \ablationTrace~fails to be steered by semantic guidance alone.
Further, the gap between \ablationMoE~and \projectname~indicates that the MoE structure is beneficial, substantially improving performance.
Lastly, \ablationMoE~slightly outperforms \ablationTrace, suggesting that the trace guidance layer can overcome a lower parameter count.

Further, the three leave-one-out ablations show that each augmentation matters, as removing any one costs 10.5\% - 13.2\% average performance drop. The trace guidance yields limited benefit unless all three augmentations are present, suggesting the augmentations are complementary rather than redundant. The ablations, however, do not show a clear rank among them.

\subsection{Pilot Studies on Noisy Semantic Guidance}
\label{sec:noise-benchmark}

Because \projectname~relies on a generalist VLM to provide semantic target points for trace generation and downstream action execution, its performance is inherently tied to the accuracy of these targets. In real-world or out-of-distribution deployments, the generalist VLM may output imperfect or noisy target predictions. To evaluate \projectname's robustness to such inaccuracies, we conducted an additional noise-injection benchmark.

\begin{figure}[htbp]
    \vspace{-5pt}
    \centering
    \includegraphics[width=0.95\linewidth]{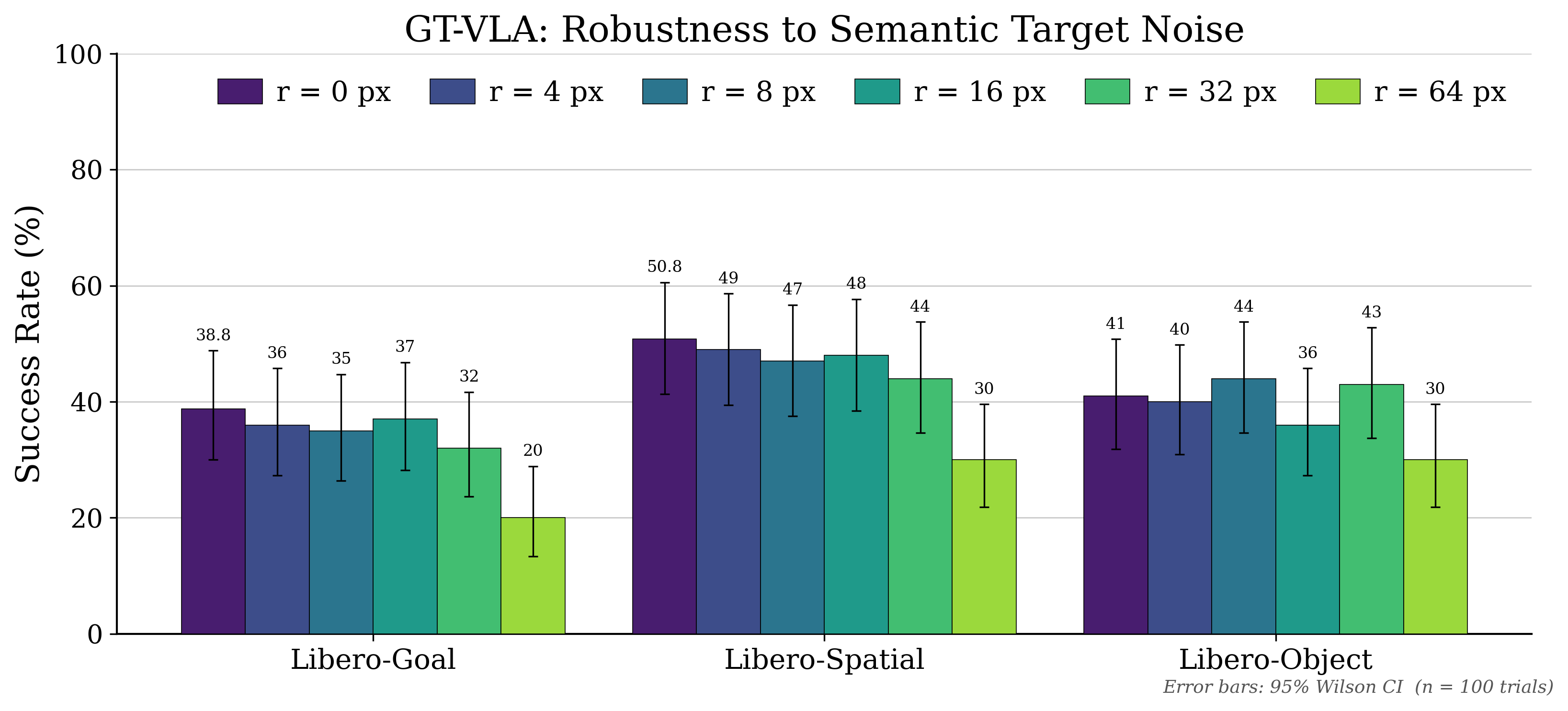}
    \includegraphics[width=0.9\linewidth]{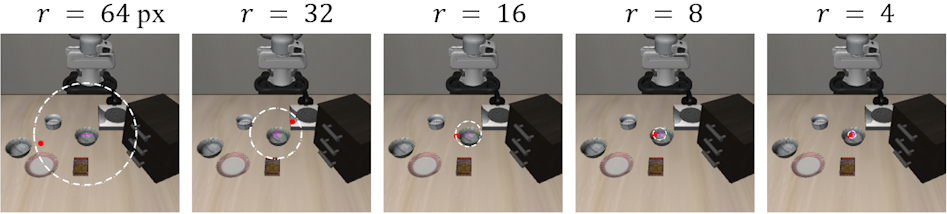}
    \caption{\textbf{Semantic target perturbations.} Top: \projectname~success rates across LIBERO suites when conditioned on noisy semantic targets. Bottom: Visual illustration of perturbations. Pink cross indicates given point, white circle indicates perturbation max. radius, red dot indicates perturbed final point. }
    \label{fig:benchmark-noisy-target}
    \vspace{-8pt}
\end{figure}

In this evaluation, we systematically perturb the semantic targets generated by the VLM. For a given original target point, we define a noise radius $r$ (in pixels) to form a bounding disk around it. We then uniformly sample a new coordinate from within this disk to serve as the perturbed target, ensuring the noise is not biased toward the original center. We test $r \in \{4, 8, 16, 32, 64\}$. Since the input observation image size is $224 \times 224$ pixels, the maximum perturbation radius of $r = 64$ encompasses over a quarter of the image width, representing severe spatial drift. Visual illustrations of these noise levels are provided in Fig.~\ref{fig:benchmark-noisy-target}.

We ran 100 trials for each noise radius across the three LIBERO out-of-distribution evaluation suites (see Sec.~\ref{sec:sim-setup}). The results, reported in Fig.~\ref{fig:benchmark-noisy-target}, demonstrate that \projectname~maintains robust performance for noise radii up to $r = 32$ pixels. Remarkably, even with this degree of perturbation, \projectname~still generally outperforms the baselines presented in Table~\ref{tab:sim-eval-results}.

The sharp performance drop observed at $r = 64$ pixels highlights a natural boundary for trace-conditioning robustness. As visualized in Fig.~\ref{fig:benchmark-noisy-target}, perturbed targets at or below $r = 32$ pixels typically remain physically grounded on or immediately adjacent to the intended object. \projectname~is robust to this localized noise because the target still provides a broadly correct semantic direction. However, at $r = 64$ pixels, the target point frequently drifts onto entirely incorrect objects or empty background space, misguiding the trace generation and significantly degrading performance.

\subsection{Pilot Studies on Generalist VLM Swap}
\label{sec:eval-vlm-swap}

In principle, \projectname's design allows it to take guidance from any high-quality off-the-shelf VLM generalist. To test this hypothesized modularity, we evaluate the impact of swapping the guidance VLM at test time to Gemini 3.6 Flash, Qwen 3.8 Max, and GPT-5.6 Terra, while freezing learned components.
As shown in Table~\ref{tab:ablation}, \projectname~remains competitive with or outperforms the baselines in Table~\ref{tab:sim-eval-results}, despite using guidance VLMs unseen during training. This also highlights the key benefit of \projectname~over related works that leverage a finetuned VLM~\cite{liu2026long}. The benefit of using an off-the-shelf generalist VLM goes beyond finetuning cost: \projectname~is not tied to a single generalist and may benefit from advances in frontier models out-of-the-box.

\subsection{Failure Analysis}
\label{sec:failure-analysis}

\begin{figure}[t]
  \centering
  \includegraphics[width=\linewidth]{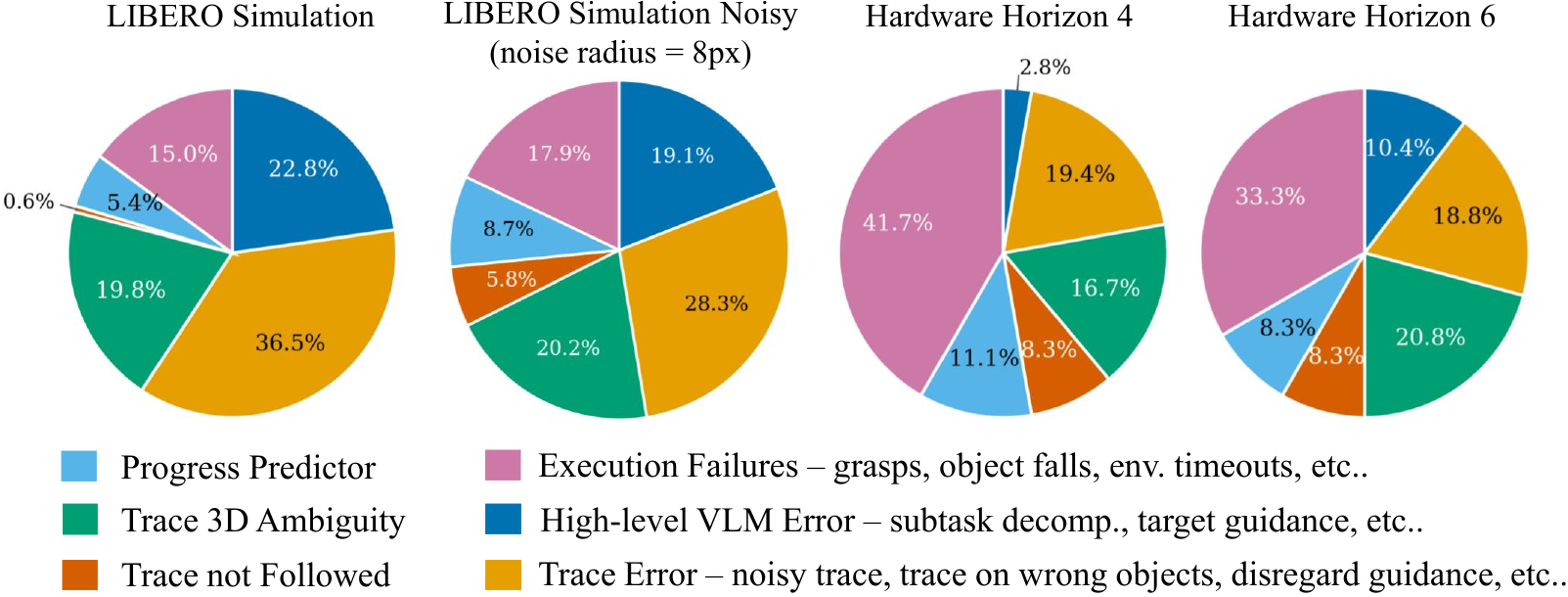}
  \caption{\projectname~failure breakdowns for simulation, hardware, and noise injections.}
  \label{fig:failure_breakdown}
\end{figure}

We assign each failed episode a single primary failure source, the
earliest pipeline stage that deviated (Fig.~\ref{fig:failure_breakdown}).
The results support three design choices. 
First, an off-the-shelf generalist suffices for sparse high-level guidance without fine-tuning. The high-level VLM and progress-prediction errors remain a minority error source across categories. 
Second, the trace-conditioned policy is
reliably steerable. Failures due to not following the guidance trace range from 0.6\% to 8.3\% only, in contrast to G-VLA, which executes memorized motions despite correct targets (Fig.~\ref{fig:running_demos}). 
Third, errors do not strongly cascade through the hierarchy and model components: perturbing the
semantic target (Sec.~\ref{sec:noise-benchmark}) and extending real-world
tasks from four to six skills leave the failure profile largely unchanged,
so noisier and longer tasks generally fail for the similar reasons rather than through
new compounding failure modes.

The remaining failures lie downstream of the generalist and point to
concrete improvements. In LIBERO, inaccurate traces are the largest source
(36.5\%). On hardware, low-level execution failures such as grasp slips
dominate (41.7\% for 4-skill tasks); these occur despite correct guidance
and would benefit directly from stronger low-level controllers. 
Finally, 2D--3D ambiguity consistently accounts for around 20\% of failures, indicating a limitation of image-space
guidance itself and motivating multi-view
or 3D traces.

\section{Conclusion}
\label{sec:conclusion}

We present \projectname, a VLA framework converting off-the-shelf VLM semantic guidance into actionable image-space traces for control. Ablations show that this intermediate trace interface surpasses direct target guidance, enhancing steerability and out-of-distribution performance. \projectname~outperforms recent baselines on unseen tasks, novel configurations, and long-horizon task composition across simulated and real environments.
In future work, we aim to incorporate 3D traces for more accurate guidance. Further, we note that our MoE closely follows~\cite{zhang2026atomicvlaunlockingpotentialatomic}, which supports continual expansion by adding a new expert and extending a new router while preserving existing experts.
We expect \projectname's design to support continual skill acquisition but this is left to future work.

\bibliographystyle{IEEEtran}
\bibliography{example}


\newpage

\appendix

\section{Appendix}

\subsection{Skill Definition}
\label{app:skill_definitions}

For the LIBERO domain, we defined 7 skills, in 5 categories:
\begin{itemize}
    \item \texttt{PICKUP\_FROM(object, location)}: Pick up an object from a surface or a container.
    \item \texttt{PLACE\_ON / PLACE\_IN(object, destination)}: Place an object in/onto a destination. These two language skills are routed to a single expert.
    \item \texttt{OPEN(container)}: Open a container (drawer or microwave).
    \item \texttt{CLOSE(container)}: Close a container (drawer or microwave).
    \item \texttt{TURN\_ON / TURN\_OFF(item)}: Turn an item on or off (stove knob). These two language skills are routed to a single expert.
\end{itemize}
This results in 5 experts for the LIBERO setting.

For our real-world experiments, we use the first 2 skill categories (\texttt{PICKUP\_FROM}, \texttt{PLACE\_ON/IN}), corresponding to two experts.

The discrete skill representation provides a compact interface between high-level reasoning and low-level execution: The skill name determines which trace expert and action expert are selected, while the natural language arguments specify the task-relevant objects or regions that the VLM backbone can understand.

\subsection{Data Annotation Pipeline}
\label{app:data_annotation}

Each demonstration consists of a high-level instruction $\ell$ and a
trajectory of robot observations and actions. We first decompose $\ell$
into an ordered sequence of skills
$P=(s^{(1)},\ldots,s^{(K)})$ using the same generalist-VLM procedure
described in Sec.~\ref{sec:skill_inference}. The demonstration is then
partitioned into the corresponding contiguous skill segments. For each segment,
the training instruction $\ell^{(i)}$ is formed by concatenating the
episode instruction with the corresponding skill $s^{(i)}$.

For every timestep within a skill segment, the ground-truth remaining
trace is obtained by projecting the future end-effector trajectory up
to the end of that skill into image space. We additionally query Gemini
3.1 Pro for one semantic target $g^{(i)}$ per skill and associate this
target with all samples in the corresponding segment. This produces the
skill-annotated demonstrations used to train the trace and action experts.

\subsection{Parameter Allocation Details}
\label{app:parameters}

\begin{table}[t]
\caption{Approximate parameter allocations for \projectname~and baselines.}
\label{tab:param-alloc}
\centering
\footnotesize
\setlength{\tabcolsep}{3pt}
\renewcommand{\arraystretch}{1.05}
\begin{tabularx}{\columnwidth}{@{}Y r@{}}
\toprule
\textbf{Part} & \textbf{Parameters} \\
\midrule
Vision encoder (Gemma)~\cite{black2024pi_0} & 400M \\
Shared VLM (Gemma)~\cite{black2024pi_0} & 2.6B \\
Action heads (shared) & 200M \\
Action head (per expert) & 230M \\
Trace heads (shared) & 72M \\
Trace head (per expert) & 57M \\
Progress predictor & 0.6M \\
\midrule
$\pi_{0.5}$ / SEAL (total) & 3.4B \\
AtomicVLA (LIBERO; 5 experts + 1 shared) & 4.6B \\
\projectname~(LIBERO; 5 experts) & 4.7B \\
AtomicVLA (real; 2 experts + 1 shared) & 3.9B \\
\projectname~(real; 2 experts) & 3.8B \\
\bottomrule
\end{tabularx}
\end{table}

Following the mixture of experts implementation from AtomicVLA~\cite{zhang2026atomicvlaunlockingpotentialatomic}, we implement the expert weights as separate dense layers for each head.
Each head ends up with some portion of shared weights (attention mechanism + AdaRMS conditioning), and some portion of weights that increases with the number of experts.
SEAL~\cite{Wu2025DoWY} does not change the $\pi_{0.5}$ architecture, only the training and inference pipelines.
AtomicVLA uses a number of experts based on skill categories, plus one additional expert as a ``shared expert".

\subsection{Approximate Inference Time Latency Breakdown}
\label{app:inference_time}

\begin{table}[t]
\caption{Inference time for different components of \projectname.}
\label{tab:runtime}
\centering
\footnotesize
\setlength{\tabcolsep}{3pt}
\renewcommand{\arraystretch}{1.08}
\begin{tabularx}{\columnwidth}{@{}Y c c@{}}
\toprule
\textbf{Component} & \textbf{Runtime} & \textbf{Frequency} \\
\midrule
VLM planning     & $\sim7\,\mathrm{s}$  & 1/task \\
VLM guidance     & $\sim5\,\mathrm{s}$  & 1/skill \\
Trace generation & $\sim52\,\mathrm{ms}$ & Every 2 action chunks \\
Action generation & $\sim74\,\mathrm{ms}$ & N/A \\
\bottomrule
\end{tabularx}
\end{table}

Table~\ref{tab:runtime} provides a rough latency breakdown of the main inference components in \projectname. 
In our experiments, the generalist VLM calls for task planning and semantic guidance were queried through OpenRouter using \texttt{gemini-3.1-pro-preview}. Therefore, the absolute latency of the VLM components can vary with the queried model, serving platform, and network conditions. In contrast, trace generation and action generation are local model forward passes and are much faster than the external VLM queries.

The generalist VLM calls dominate the wall-clock latency, but they are invoked at relatively low frequency: planning is performed once per task, and semantic guidance is queried once per skill. The trace and action modules are invoked during closed-loop execution, and their runtimes are on the order of tens of milliseconds. If lower latency is required, one practical option is to replace the generalist VLM with a faster model or a lower-latency serving backend.
This may trade semantic accuracy for inference speed. Nonetheless, the robustness study in Sec.~\ref{sec:noise-benchmark} suggests that \projectname~can tolerate a moderate amount of noise in the predicted semantic target.

\subsection{LoHo-Manip Reproduction}
\label{app:loho-reproduction}

LoHo-Manip~\cite{liu2026long} is closely related to \projectname, as it also uses image-space trace guidance for long-horizon manipulation. Since no official implementation was available at the time of our evaluation, we provide a best-effort reproduction based on the architecture and training procedure described in the paper. We evaluate it under the same out-of-distribution protocol as Table~I: the model is trained on LIBERO-10 and LIBERO-90, and evaluated without per-suite fine-tuning on LIBERO-Goal, LIBERO-Spatial, and LIBERO-Object.

For the high-level manager, we initialize from \texttt{Qwen3-VL-4B-Instruct} and fine-tune for three epochs, by which point the validation loss had plateaued. The vision encoder is frozen, while the language model and multimodal projector are fully fine-tuned using next-token cross-entropy on the generated response. Given the task instruction, current image observation, and a textual memory of completed subtasks, the manager predicts both the completed/remaining subtask split and a 20-waypoint 2D trace, with coordinates normalized to $[0,1000]$, following the output representation in LoHo-Manip~\cite{liu2026long}. 
Based on validation loss, we selected the epoch-1.5 and epoch-2 manager
checkpoints for downstream evaluation and report the epoch-1.5 checkpoint,
which achieves the stronger aggregate performance across the three suites.

The $\pi_{0.5}$-based low-level executor is trained for 100k steps with batch size 64, matching the training budget used for the baselines in Table~I. We also apply the same trace augmentations used in \projectname~during executor training since we found that to be helpful from our ablation studies (Sec.~\ref{sec:ablation}). Both the manager and executor are trained using 4 GH200 GPUs.

\begin{table}[t]
\caption{Supplementary comparison with LoHo-Manip under the same LIBERO OOD protocol as Table~I. Results use 50 trials $\times$ 10 tasks per suite (\%). Parentheses show 95\% Wilson confidence intervals. The \projectname~row is repeated from Table~I for direct comparison.}
\label{tab:loho-reproduction}
\centering
\scriptsize
\setlength{\tabcolsep}{2.5pt}
\renewcommand{\arraystretch}{1.08}
\resizebox{\columnwidth}{!}{%
\begin{tabular}{lcccc}
\toprule
\textbf{Model} &
\textbf{Goal} &
\textbf{Spatial} &
\textbf{Object} &
\textbf{Avg.} \\
\midrule
LoHo-Manip (reprod.) &
30.8 (26.9--35.0) &
23.2 (19.7--27.1) &
12.8 (10.2--16.0) &
22.3 \\
\projectname~(ours) &
38.8 (34.5--43.2) &
50.8 (46.3--55.3) &
41.8 (37.4--46.3) &
43.8 \\
\bottomrule
\end{tabular}%
}
\end{table}

As shown in Table~\ref{tab:loho-reproduction}, the reproduced LoHo-Manip achieves an average success rate of 22.3\%, compared with 43.8\% for \projectname. The difference is relatively smaller on LIBERO-Goal (30.8\% vs.~38.8\%) and substantially larger on LIBERO-Spatial (23.2\% vs.~50.8\%) and LIBERO-Object (12.8\% vs.~41.8\%). While this comparison is necessarily limited by the absence of an official LoHo-Manip implementation, the results provide an additional comparison against a closely related trace-conditioned hierarchical approach under our OOD evaluation protocol.

\subsection{Real-world Experiment Training and Evaluation Tasks}
\label{app:real-world-tasks}

\subsubsection{Training Task Setup}
The training dataset is split across 4 settings, with different objects on the table, described in Table~\ref{tab:training_settings}.
For each setting, 3 high-level tasks were demonstrated, described in Table~\ref{tab:training_tasks}.
For each high-level task, the table was shuffled 5 times to mix up object positions, then the task was demonstrated 5 times for each shuffle.
This gives $5 \times 5 \times 3 \times 4 = 300$ demonstrations nominally, with some human errors resulting in 299 total episodes.

\begin{table}[t]
\caption{Real-world training-data settings.}
\label{tab:training_settings}
\centering
\footnotesize
\setlength{\tabcolsep}{3pt}
\renewcommand{\arraystretch}{1.10}
\begin{tabularx}{\columnwidth}{@{}l Y@{}}
\toprule
\textbf{Setting} & \textbf{Objects} \\
\midrule
\textbf{table1} &
orange plate, blue plate, basket, eggplant, corn, carrot,
green bell pepper \\

\textbf{table2} &
white plate, green plate, orange plate, eggplant, corn, carrot,
green bell pepper, asparagus \\

\textbf{table3} &
blue plate, green plate, red bell pepper, green bell pepper, corn \\

\textbf{table4} &
white plate, basket, green bell pepper, carrot, eggplant, asparagus \\
\bottomrule
\end{tabularx}
\end{table}

\begin{table*}[!t]
\caption{Real-world training tasks. Objects are initially placed on the table unless otherwise specified; ``--'' denotes no additional initial-state constraint.}
\label{tab:training_tasks}
\centering
\small
\setlength{\tabcolsep}{5pt}
\renewcommand{\arraystretch}{1.10}

\begin{tabularx}{\textwidth}{
    @{}
    >{\centering\arraybackslash}p{0.45cm}
    Y
    >{\centering\arraybackslash}p{1.25cm}
    >{\raggedright\arraybackslash}p{3.5cm}
    @{}
}
\toprule
\textbf{ID} & \textbf{Task Instruction} & \textbf{Setting} & \textbf{Initial State} \\
\midrule

0 &
move the basket onto the blue plate and put the green bell pepper in the basket &
table1 & -- \\

1 &
move the orange plate onto the basket &
table1 & -- \\

2 &
move the eggplant into the basket and place the blue plate on the basket &
table1 & eggplant on blue plate \\

\addlinespace[3pt]

3 &
move the green plate onto the orange plate and move the asparagus and corn onto the white plate &
table2 & -- \\

4 &
put the carrot onto the green plate &
table2 & green bell pepper on green plate \\

5 &
move the eggplant and carrot onto the white plate &
table2 & corn on white plate \\

\addlinespace[3pt]

6 &
put the blue plate on the green plate and place the corn and red bell pepper on the blue plate &
table3 & -- \\

7 &
place the green bell pepper on the green plate &
table3 & green plate on blue plate \\

8 &
put the red bell pepper onto the blue plate and move the green bell pepper and corn to the green plate &
table3 & green bell pepper on blue plate \\

\addlinespace[3pt]

9 &
place the eggplant, asparagus, and carrot onto the white plate &
table4 & -- \\

10 &
move the carrot, eggplant, and green bell pepper into the basket &
table4 & -- \\

11 &
place the eggplant and green bell pepper onto the white plate &
table4 & -- \\

\bottomrule
\end{tabularx}
\end{table*}

\subsubsection{Evaluation Task Setup}
\begin{figure*}[!t]
    \centering
    \includegraphics[width=\linewidth]{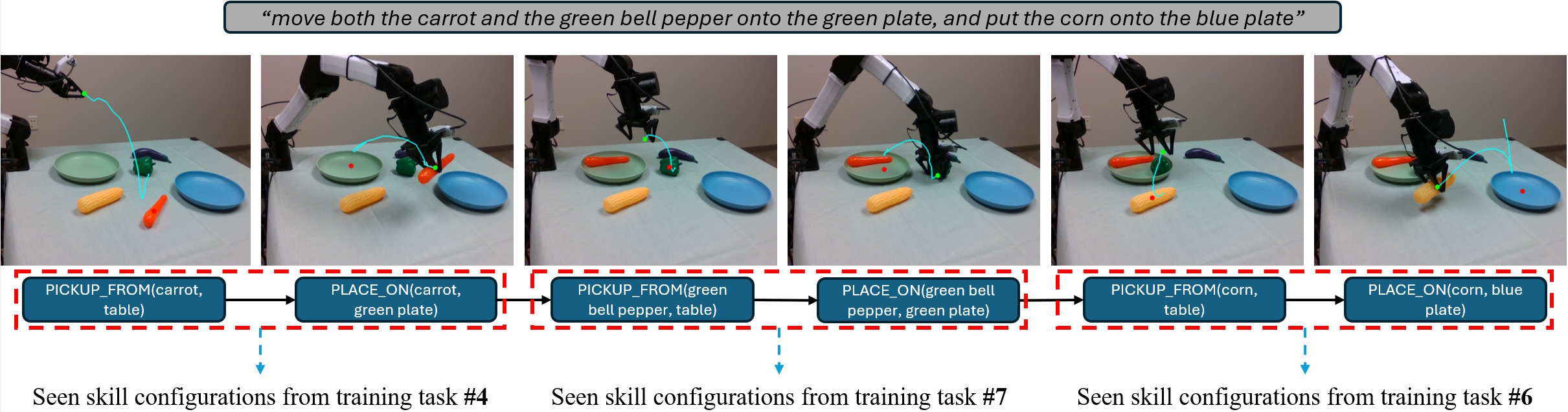}
    \caption{Visual illustration of a task setup from the \textbf{composition-6} test suite. The seen skill configurations identified in the figure can be found in Table~\ref{tab:training_tasks}.}
    \label{fig:comp-setup-vis}
\end{figure*}

The \textbf{composition} suites evaluate the model's ability to chain familiar skills into novel sequences. While all individual skill configurations are encountered during training, they are drawn from different training tasks to form completely unseen combinations. For example, Fig.~\ref{fig:comp-setup-vis} illustrates a \textbf{composition-6} task, which requires the execution of six previously learned skills in a novel sequence.

\begin{figure*}[!t]
    \centering
    \includegraphics[width=\linewidth]{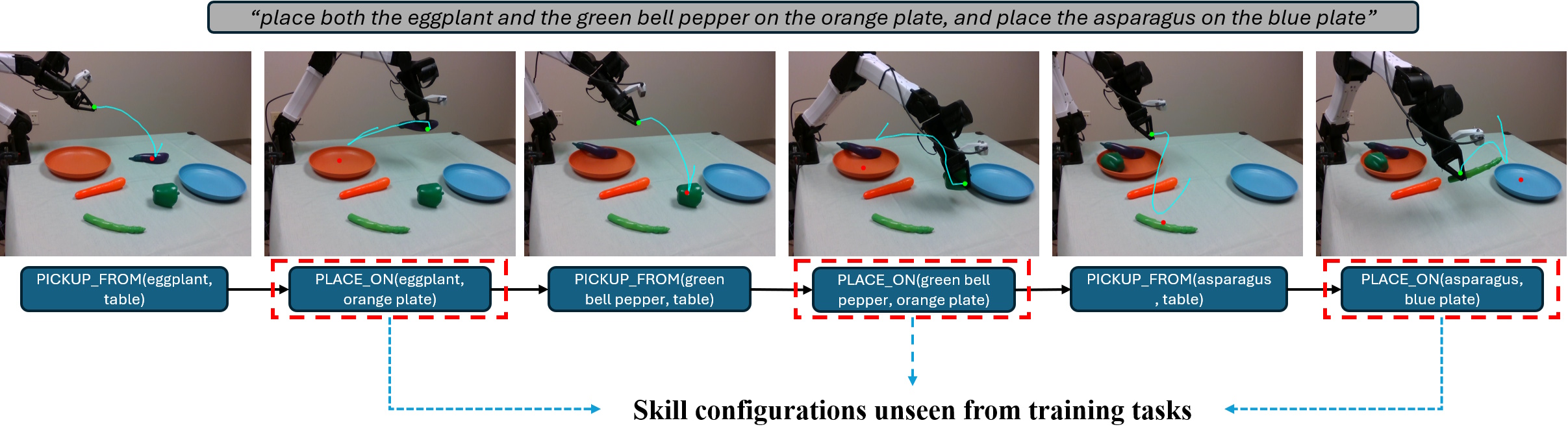}
    \caption{Visual illustration of a task setup from the \textbf{OOD-6} test suite. The boxed skill configurations are unseen from the training dataset.}
    \label{fig:ood-setup-vis}
\end{figure*}

The \textbf{OOD} suites evaluate the model on tasks containing novel skill configurations not present in the training data. For example, Fig.~\ref{fig:ood-setup-vis} illustrates an \textbf{OOD-6} task requiring a sequence of six skills, where the configurations boxed in red are unseen skill configurations.

We evaluate each task over five episodes. Each episode initializes with random object placements and may include random visual distractors. To ensure fair comparisons across all methods, the object layouts are kept visually identical for each corresponding episode.

In the following, we present all \textbf{composition} and \textbf{OOD} task instructions. Visual demonstrations of select tasks can be found in the supplemental video. 

\textbf{Composition-4}
\begin{enumerate}
    \item \textit{place the corn on the blue plate and the carrot on the white plate}  
    
    \item \textit{put the eggplant into the basket and put the basket onto the blue plate} 
    
    \item \textit{move the green bell pepper into the basket, and put the blue plate onto the basket}  
    
    \item \textit{place the red bell pepper onto the blue plate and put the corn on the white plate}  
    
    \item \textit{move the carrot onto the green plate and place the green bell pepper onto the green plate}  
    
    \item \textit{place the carrot into the basket and place the red bell pepper onto the blue plate} 
    
    \item \textit{put the eggplant on the white plate and put the carrot on the green plate} 
    
    \item \textit{move the green bell pepper onto the green plate and place the asparagus onto the white plate} 
    
    \item \textit{put the asparagus on the white plate and place the carrot into the basket} 
    
    \item \textit{place the green bell pepper into the basket and put the red bell pepper onto the blue plate} 
\end{enumerate}

\textbf{Composition-6}
\begin{enumerate}
    \item \textit{put the corn on the blue plate, and put both the carrot and the asparagus onto the white plate}  
    
    \item \textit{put the eggplant and the green bell pepper into the basket, and put the basket onto the blue plate}  
    
    \item \textit{put the eggplant and the green bell pepper in the basket, and place the blue plate onto the basket}  

    \item \textit{place the eggplant and the corn on the white plate, and put the red bell pepper onto the blue plate}  

    \item \textit{move both the carrot and the green bell pepper onto the green plate, and put the corn onto the blue plate}  

    \item \textit{place the carrot into the basket, and place the corn and the red bell pepper onto the blue plate}  

    \item \textit{put the corn and the carrot on the green plate, and place the eggplant onto the white plate}  

    \item \textit{move the green bell pepper onto the green plate, and place both the carrot and the asparagus onto the white plate}  

    \item \textit{move both the green bell pepper and the asparagus onto the white plate, and put the carrot into the basket}  

    \item \textit{put the corn and the red bell pepper on the blue plate, and place the green bell pepper into the basket}  
\end{enumerate}

\textbf{OOD-2}
\begin{enumerate}
    \item \textit{place the blue plate on the orange plate}  
    
    \item \textit{put the eggplant on the green plate}  
    
    \item \textit{place the green bell pepper on the orange plate}  

    \item \textit{put the asparagus onto the green plate}  

    \item \textit{place the corn on the orange plate}  

    \item \textit{place the carrot onto the orange plate}  

    \item \textit{place the red bell pepper onto the orange plate}  

    \item \textit{put the orange plate on the white plate}  

    \item \textit{place the green plate on the blue plate}  

    \item \textit{put the eggplant onto the blue plate}  
\end{enumerate}

\textbf{OOD-4}
\begin{enumerate}
    \item \textit{place the carrot onto the blue plate and put the corn into the basket}  
    
    \item \textit{put both the green bell pepper and the eggplant on the blue plate}  
    
    \item \textit{place the eggplant on the green plate and put the asparagus in the basket}  

    \item \textit{place the asparagus on the green plate and put the corn on the orange plate}  

    \item \textit{place the red bell pepper on the green plate and put the carrot on the orange plate}  

    \item \textit{place the green plate on the white plate and place the red bell pepper on the green plate}  

    \item \textit{place the orange plate on the green plate and put the blue plate on the orange plate}  

    \item \textit{put both the green bell pepper and the carrot on the orange plate}  

    \item \textit{place the red bell pepper into the basket, and put the corn into the basket}  

    \item \textit{place the eggplant on the orange plate, and put the red bell pepper on the white plate}  
\end{enumerate}

\textbf{OOD-6}
\begin{enumerate}
    \item \textit{put both the eggplant and the green bell pepper on the blue plate, and place the corn into the basket}  
    
    \item \textit{put both the eggplant and the red bell pepper on the green plate, and place the asparagus into the basket}  
    
    \item \textit{put the carrot on the blue plate and place both the corn and the eggplant onto the orange plate}  

    \item \textit{place both the eggplant and the green bell pepper on the orange plate, and place the asparagus on the blue plate}  

    \item \textit{place both the red bell pepper and the corn into the basket, and place the carrot onto the blue plate}  

    \item \textit{put the corn on the orange plate, and put the eggplant onto the green plate, and place the green bell pepper on the orange plate}  

    \item \textit{place the asparagus and the carrot on the blue plate, and place the eggplant on the green plate}  

    \item \textit{put the orange plate onto the blue plate, and place both the asparagus and the corn on the orange plate}  

    \item \textit{place the corn on the orange plate, and put both the eggplant and the carrot on the blue plate}  

    \item \textit{place the red bell pepper on the white plate, and place both the asparagus and the eggplant onto the green plate}  
\end{enumerate}

\subsection{Prompt Templates}
\label{app:prompts}

\lstdefinestyle{promptstyle}{
  basicstyle=\ttfamily\fontsize{6.2pt}{6.6pt}\selectfont,
  breaklines=true,
  breakatwhitespace=false,
  columns=fullflexible,
  frame=single,
  framerule=0.4pt,
  xleftmargin=0.3em,
  xrightmargin=0.1em,
  aboveskip=0.3em,
  belowskip=0.3em,
  lineskip=-0.4pt,
  showstringspaces=false
}

\subsubsection{Skill Planning Prompt Template}

As described in Sec.~\ref{sec:skill_inference}, given a task instruction $\ell$ and the current observation $o_t$, we query an off-the-shelf generalist VLM $\Phi$ to produce a skill plan using the following prompt template, along with the side-view image.

\begin{lstlisting}[style=promptstyle]
You are planning a tabletop robot manipulation task.
You are given:
- the natural-language task instruction for this episode
- the observation of the INITIAL scene

Your task:
1. Decide the ordered sequence of atomic skills that, executed in order, will accomplish
   the instruction from the visible initial state.
2. Use ONLY the allowed skills below and follow the syntax exactly.

Task instruction: {task_instruction}

{SKILL_DEFINITIONS_TEXT}

Planning rules:
- The first skill must be PICKUP_FROM - at the start the gripper holds nothing, so it
  cannot PLACE_ON or PLACE_IN.
- A PLACE_ON / PLACE_IN of an object must come right after the PICKUP_FROM of that object
  (pick, then place, then pick the next object).
- Object descriptions must be short, specific, contain no commas, and reuse the
  appearance / color cues from the instruction (e.g. "green bell pepper", "blue plate").
- Keep the plan as short as possible while still completing the instruction.

Output format:
- Return JSON only.
- "plan" must be a single numbered string of the exact form:
  "1. PICKUP_FROM(green bell pepper, table) 2. PLACE_ON(green bell pepper, blue plate)"

Example:
{{"reasoning": "the carrot is on the table and must go on the blue plate; the corn goes in the basket",
  "plan": "1. PICKUP_FROM(carrot, table) 2. PLACE_ON(carrot, blue plate) 3. PICKUP_FROM(corn, table) 4. PLACE_IN(corn, basket)"}}
\end{lstlisting}

\subsubsection{Target Acquisition Prompt Template}

\projectname~queries the generalist model for a semantic target at the start of each skill. Given the current skill, the generalist model provides a target in the normalized image coordinates. We use the following prompt for target acquisition. 

\begin{lstlisting}[style=promptstyle]
You are annotating a tabletop robot manipulation skill from the observation of the current scene.

Episode instruction: {task_instruction}
Current skill (the next one to execute):
- skill: {skill_text}
- sent image size: width={image_width}, height={image_height}

Semantic hints:
1. For PICKUP_FROM, point at the center of the visible body of the object to be picked up.
2. For PLACE_ON / PLACE_IN, point at the receiving surface or the inside of the receiving container.

Coordinate rules:
- Return coordinates on a fixed output grid width={coordinate_grid}, height={coordinate_grid}.
- The full image spans x=0..{coordinate_grid} left-to-right and y=0..{coordinate_grid} top-to-bottom.
- Return point_2d in row-column order: [y, x], not [x, y].
- The point must be on the visible object (or contact location), strictly inside the image bounds.
- "label" should be the literal string "semantic_target".
- Return JSON only - no markdown, prose, or code fences outside the JSON object.

Output shape:
{{"status": "OK",
  "reasoning": "brief visual justification",
  "label": "semantic_target",
  "point_2d": [y, x]}}
\end{lstlisting}

\end{document}